\documentclass[epsf, twocolumn]{IEEEtran}
\usepackage{graphicx}

\newtheorem{remark}{Remark}[section]

\begin{document}

\title{A New Analytical Radial Distortion Model for Camera Calibration
\thanks{
   The authors are with the Center for Self-Organizing and Intelligent Systems (CSOIS), Dept. of Electrical and Computer Engineering,  4160 Old Main Hill, Utah State University, Logan, UT 84322-4160, USA.
    This work is supported in part by U.S. Army Automotive and Armaments Command (TACOM)
 Intelligent Mobility Program (agreement no. DAAE07-95-3-0023). Corresponding author: Dr YangQuan Chen. E-mail:  \texttt{yqchen@ieee.org}; Tel. 01-435-7970148; Fax: 01-435-7972003. URL: \texttt{http://www.csois.usu.edu/people/yqchen}. }
}

\author{\vskip 1em
Lili Ma, YangQuan Chen, and Kevin L. Moore \vskip 1em
}

\maketitle{}

\begin{abstract}
Common approach to radial distortion is by the means of polynomial approximation, which introduces distortion-specific parameters into the camera model and requires estimation of these distortion parameters. The task of estimating radial distortion is to find a radial distortion model that allows easy undistortion as well as satisfactory accuracy. This paper presents a new radial distortion model with an  easy analytical undistortion formula, which also belongs to the polynomial approximation category. Experimental results are presented to show that with this radial distortion model, satisfactory accuracy is achieved.
\end{abstract}

\thispagestyle{empty}
\section{Introduction}

Cameras are widely used in many engineering automation processes from visual monitoring, visual metrology to real time visual servoing or visual following. 
We will focus on a new polynomial radial distortion model which introduces a quadratic term yet having an analytical undistortion formula.

\subsection{Camera Calibration}
Camera calibration is to estimate a set of parameters that describes the camera's imaging process. With this set of parameters, a perspective projection matrix can directly link a point in the 3-D world reference frame to its projection (undistorted) on the image plane by:
\begin{equation}
\label{eqn: projection matrix}
\lambda \left [\matrix{u \cr v \cr 1} \right ] 
= {\bf A} \, \left[{\bf R} \mid {\bf t}\right] \left [\matrix{X^w \cr Y^w \cr Z^w \cr 1} \right ] 
= \left[\matrix{\alpha & \gamma &u_0\cr 0 & \beta & v_0\cr 0 & 0 &1}\right] \left [\matrix{X^c \cr Y^c \cr Z^c} \right ],
\end{equation}
where $(u,v)$ is the distortion-free image point on the image plane; the matrix $\bf A$ fully depends on the camera's 5 intrinsic parameters $(\alpha, \gamma, \beta, u_0, v_0)$ with $(\alpha, \beta)$ being two scalars in the two image axes, $(u_0, v_0)$ the coordinates of the principal point, and $\gamma$ describing the skewness of the two image axes; $[X^c, Y^c, Z^c]^T$ denotes a point in the camera frame which is related to the corresponding point $[X^w, Y^w, Z^w]^T$ in the world reference frame by $P^c = {\bf R} P^w + \bf t$ with $({\bf R}, {\bf t})$ being the rotation matrix and the translation vector. For a variety of computer vision applications where camera is used as a sensor in the system, the camera is always assumed fully calibrated beforehand. 

The early works on precise camera calibration, starting in the photogrammetry community, use a 3-D calibration object whose geometry in the 3-D space is required to be known with a very good precision. However, since these approaches require an expensive calibration apparatus, camera calibration is prevented from being carried out broadly. Aiming at the general public, the camera calibration method proposed in \cite{zhang99calibrationinpaper} focuses on desktop vision system and uses 2-D metric information. The key feature of the calibration method in \cite{zhang99calibrationinpaper} is that it only requires the camera to observe a planar pattern at a few (at least 3, if both the intrinsic and the extrinsic parameters are to be estimated uniquely) different orientations without knowing the motion of the camera or the calibration object. Due to the above flexibility, the calibration method in \cite{zhang99calibrationinpaper} is used in this work where the detailed procedures are summarized as: 1) estimation of intrinsic parameters, 2) estimation of extrinsic parameters, 3) estimation of distortion coefficients, and 4) nonlinear optimization.

\subsection{Radial Distortion}
In equation (\ref{eqn: projection matrix}), $(u,v)$ is not the actually observed image point since virtually all imaging devices introduce certain amount of nonlinear distortions. Among the nonlinear distortions, radial distortion, which is performed along the radial direction from the center of distortion, is the most severe part \cite {OlivierF01Straight,tsai87AVersatile}. The radial distortion causes an inward or outward displacement of a given image point from its ideal location. The negative radial displacement of the image points is referred to as the barrel distortion, while the positive radial displacement is referred to as the pincushion distortion \cite{Juyang92distortionmodel}. The removal or alleviation of the radial distortion is commonly performed by first applying a parametric radial distortion model, estimating the distortion coefficients, and then correcting the distortion. 

Lens distortion is very important for accurate 3-D measurement \cite{Tsai88Techniques}. 
Let $(u_d, v_d)$ be the actually observed image point and assume that the center of distortion is at the principal point. The relationship between the undistorted and the distorted radial distances is given by 
\begin{equation}
r_d = r + \delta_r,
\end{equation}
where $r_d$ is the distorted radial distance and $\delta_r$ the radial distortion (some other variables used throughout this paper are listed in Table \ref{table: variables used}). 

\begin{table}[htb]
\centering
\caption{List of Variables}
\label{table: variables used}
\renewcommand{\arraystretch}{1.3}
\vspace{-2mm}
{
{\begin {tabular}{|c|l|}\hline
{\bf Variable} & {\bf Description} \\[1ex]\hline
$(u_d, \, v_d)$              & Distorted image point in pixel\\\hline
$(u, \, v)$                  & Distortion-free image point in pixel\\\hline
$(x_d, \, y_d)$              & $[x_d, \, y_d, \, 1]^T = A^{-1} [u_d, \, v_d, \, 1]^T$\\\hline
$(x, \, y)$                  & $[x,\, y,\, 1]^T = A^{-1} [u,\, v,\, 1]^T$ \\\hline
$r_d$                        & $r_d^2 = x_d^2 + y_d^2$ \\\hline
$r$                          & $r^2 = x^2 + y^2$ \\\hline
$\bf k$                      & Radial distortion coefficients \\\hline
\end {tabular}}}   
\end{table}

Most of the existing works on the radial distortion models can be traced back to an early study in photogrammetry \cite{Photogrammetry80} where the radial distortion is governed by the following polynomial equation \cite{zhang99calibrationinpaper,Photogrammetry80,Heikkil97fourstepcameracalibration,Janne96Calibration}:
\begin{equation}
\label{eqn: general polynomial}
r_d = r \, f(r) = r \, (1 + k_1 r^2 + k_2 r^4 + k_3 r^6+ \cdots),
\end{equation}
where $k_1, k_2, k_3, \ldots$ are the distortion coefficients. It follows that 
\begin{equation}
\label{eqn: radial xdyd from xy}
x_d = x \, f(r), \;\;\; y_d = y \, f(r),
\end{equation}
which is equivalent to 
\begin{eqnarray}
\label{eqn: radial (ud,vd) (u,v) relation}
\left\{\hspace{-1mm}
\begin{array}{c}
u_d - u_0 = (u-u_0) \, f(r) \\
v_d - v_0 = (v-v_0) \, f(r)
\end{array}\right.\hspace{-1 mm}.
\end{eqnarray}
This is because
\begin{eqnarray*}
u_d &=& \alpha \, x_d + \gamma \, y_d + u_0, 		\\
    &=& \alpha \, x f(r) + \gamma \, y f(r) + u_0 	\\
    &=& (u-u_0) \, f(r) + u_0 				\\
v_d &=& \beta \, y_d  + v_0.					\\
    &=& (v-v_0) \, f(r) + v_0					
\end{eqnarray*}

For the polynomial radial distortion model (\ref{eqn: general polynomial}) and its variations, the distortion is especially dominated by the first term and it has also been found that too high an order may cause numerical instability \cite{tsai87AVersatile,zhang99calibrationinpaper,GWei94Implicit}. In this paper, at most two terms of radial distortion are considered. When using two coefficients, the relationship between the distorted and the undistorted radial distances becomes \cite{zhang99calibrationinpaper}
\begin{eqnarray}
\label{eqn: radial distortion order 2 4}
r_d = r \, (1 + k_1 \, r^2 + k_2 \, r^4).
\end{eqnarray}
The inverse of the polynomial function in (\ref{eqn: radial distortion order 2 4}) is difficult to perform analytically but can be obtained numerically via an iterative scheme. In \cite{Undistortionchapter}, for practical purpose, only one distortion coefficient $k_1$ is used. 



For a specific radial distortion model, the estimation of distortion coefficients and the correction of radial distortion can be done by correspondences between feature points (such as corners \cite{zhang99calibrationinpaper} and circles \cite{Janne00Geometric}), image registration \cite{Torn02Correcting}, the plumb-line algorithm \cite{Brown71CloseRange}, and the blind removal technique \cite{Hany01blindremoval} that exploits the fact that lens distortion introduces specific higher-order correlations in the frequency domain. However, this paper mainly focuses on the radial distortion models, advantages and disadvantages of the above four calibration methods are not further discussed. 

In this work, a new radial distortion model is proposed that belongs to the polynomial approximation category. 
To compare the performance of different distortion models, final value of optimization function $J$, which is defined to be \cite{zhang99calibrationinpaper}:
\begin{equation}
\label{eqn: objective function}
J = \sum_{i=1}^N \sum_{j=1}^n \|m_{i j}-\hat m({\bf A}, {\bf k}, {\bf R}_i, {\bf t}_i, M_j) \|^2,
\end{equation}
is used, where $\hat m({\bf A}, {\bf k}, {\bf R}_i, {\bf t}_i, M_j)$ is the projection of point $M_j$ in the $i^{th}$ image using the estimated parameters; $\bf k$ denotes the distortion coefficients; $M_j$ is the $j^{th}$  3-D point in the world frame with $Z^w = 0$; $n$ is the number of feature points in the coplanar calibration object; and $N$ is the number of images taken for calibration.
In \cite{zhang99calibrationinpaper}, the estimation of radial distortion is done after having estimated the intrinsic and the extrinsic parameters and just before the nonlinear optimization step. So, for different radial distortion models, we can reuse the estimated intrinsic and extrinsic parameters.

The rest of the paper is organized as follows. Sec.~\ref{sec: Polynomial Radial Distortion Models} describes the new radial distortion model and its inverse analytical formula. Experimental results and comparison with the existing polynomial models are presented in Sec.~\ref{sec: Comparison Results}. Finally, some concluding remarks are given in Sec.~\ref{sec: Conclusion}.

\thispagestyle{empty}
\section{The New Radial Distortion Model}
\label{sec: Polynomial Radial Distortion Models}

\subsection{Model}
\label{sec: new model}

The conventional radial distortion model (\ref{eqn: radial distortion order 2 4}) with 2 parameters does not have an exact inverse, though there are ways to approximate it without iterations, such as the model described in \cite{Janne00Geometric}, where $r$ can be calculated from $r_d$ by
\begin{equation}
\label{eqn: undistortion approximation}
r =  r_d \, ( 1 - k_1 \, r_d^2 - k_2 \, r_d^4).
\end{equation}
The fitting results given by the above model can be satisfactory when the distortion coefficients are small values. However,   equation (\ref{eqn: undistortion approximation}) introduces another source of error that will inevitably degrade the calibration accuracy. Due to this reason, an analytical inverse function that has the advantage of giving the exact undistortion solution is one of the main focus of this work. 

To overcome the shortcoming of no analytical undistortion formula but still preserving a comparable accuracy, the new radial distortion model is proposed as \cite{LiliISIC03Flex}:
\begin{equation}
\label{eqn: new distortion model}
r_d = r \, f(r) = r \, (1 + k_1\, r + k_2\, r^2),
\end{equation}
which has the following three properties:
\begin{itemize}
\item[{\rm \bf 1)}] This function is radially symmetric around the center of distortion (which is assumed to be at the principal point $(u_0, v_0)$ for our discussion) and it is expressible in terms of radius $r$ only;
\item[{\rm \bf 2)}] This function is continuous, hence $r_d = 0$ iff $r = 0$;
\item[{\rm \bf 3)}] The resultant approximation of $x_d$ is an odd function of $x$, as can be seen next.
\end{itemize}
Introducing a quadratic term $k_1 \, r^2$ in (\ref{eqn: new distortion model}), the new distortion model still approximates the radial distortion, since the distortion is in the radial sense. 

From (\ref{eqn: new distortion model}), we have
\begin{equation}
\left \{\hspace{-1mm}
\begin{array}{l}
x_d = x \, f(r) = x \, (1 + k_1 r + k_2 r^2)  \\
y_d = y \, f(r) = y \, (1 + k_1 r + k_2 r^2)
\end{array}\right. .
\end{equation}
It is obvious that $x_d = 0$ iff $x = 0$. When $x_d \ne 0$, by letting $c = y_d/x_d = y /x$, we have $y = cx$ where $c$ is a constant. Substituting $y = cx$ into the above equation gives
\begin{eqnarray}
\label{eqn: distortion model 3}
x_d &=& x \, \left[ 1+k_1 \sqrt{x^2 + c^2x^2} + k_2(x^2 + c^2x^2)\right] \nonumber\\
   &=& x \, \left[ 1+k_1 \sqrt{1 + c^2} \, {\tt sgn}(x) x + k_2(1 + c^2)x^2 \right] \nonumber\\
   &=& x + k_1 \sqrt{1 + c^2} \, {\tt sgn}(x) \, x^2 + k_2(1 + c^2) \, x^3,
\end{eqnarray}
where ${\tt sgn}(x)$ gives the sign of $x$ and $x_d$ is an odd function of $x$. 

The well-known radial distortion model (\ref{eqn: general polynomial}) that describes the laws governing the radial distortion does not involve a quadratic term. 
Thus, it might be unexpected to add one. However, when interpreting from the relationship between $(x_d,y_d)$ and $(x,y)$ in the camera frame as in equation (\ref{eqn: distortion model 3}), the radial distortion function is to approximate the $x_d \leftrightarrow x$ relationship which is intuitively an odd function. Adding a quadratic term to $\delta_r$ does not alter this fact. Furthermore, introducing quadratic terms to $\delta_r$ broadens the choice of radial distortion functions.

\begin{remark}
The radial distortion models discussed in this paper belong to the category of {\underline U}ndistorted-{\underline D}istorted model, while the {\underline D}istorted-{\underline U}ndistorted model also exists in the literature to correct the distortion \cite{Toru02Unified}. The new radial distortion model can be applied to the D-U formulation simply by defining
\begin{equation}
\label{eqn: D-U model in xy}
r = r_d \, (1 + {\tilde k}_1 \, r_d + {\tilde k}_2 \, r_d^2).
\end{equation}
Consistent results and improvement can be achieved in the above D-U formulation. 
\end{remark}

\subsection{Radial Undistortion of The New Model}
\label{sec: undistortion}

From (\ref{eqn: new distortion model}), we have
\begin{displaymath}
r^3 + a \, r^2 + b \, r + c = 0,
\end{displaymath}
with $a = k_1/k_2, b = 1/k_2$, and $c = -r_d/k_2$. Let ${\bar r} = r - a/3$, the above equation becomes
\begin{displaymath}
{\bar r}^3 + p \, {\bar r} + q = 0,
\end{displaymath}
where $p = b - a^2/3, \; q = 2a^3/27 - a b/3 + c$. Let $\Delta = (\frac{q}{2})^2 + (\frac{p}{3})^3$. If $\Delta > 0$, there is only one solution; if $\Delta = 0$, then $r = 0$ , which occurs when $\delta_r = 0$; if $\Delta < 0$, then there are three solutions. In general, the middle one is what we need, since the first root is at a negative radius and the third lies beyond the positive turning point \cite{zhang96OnThe,Ben02PhD}. 
After $r$ is determined, $(u,v)$ can be calculated from (\ref{eqn: radial (ud,vd) (u,v) relation}) uniquely. 

The purpose of this work is to show that by adding a quadratic term to $\delta_r$, the resultant new model achieves the following properties: 
\begin{itemize}
\item [\bf 1)] Given $r_d$ and the distortion coefficients, the solution of $r$ from $r_d$ has closed-form solution; 
\item [\bf 2)] It approximates the commonly used distortion model (\ref{eqn: radial distortion order 2 4}) with higher accuracy than $f(r) = 1 + k \, r^2$ based on the final value of the optimization function $J$ in (\ref{eqn: objective function}), as will be presented in Sec.~\ref{sec: Comparison Results}.
\end{itemize}

\thispagestyle{empty}
\section{Experimental Results and Comparisons}
\label{sec: Comparison Results}


In this section, the performance comparison of our new radial distortion model with two other existing models is presented based on the final value of objective function $J$ in (\ref{eqn: objective function}) after nonlinear optimization by the Matlab function {\tt fminunc}, since common approach to camera calibration is to perform a full-scale nonlinear optimization for all parameters. The three different distortion models for comparison are:
\begin{eqnarray*}
{\tt distortion \;\; model_1:} && f(r) = 1 + k_1\, r^2 + k_2\, r^4,      \\
{\tt distortion \;\; model_2:} && f(r) = 1 + k_1\, r^2,                  \\
{\tt distortion \;\; model_3:} && f(r) = 1 + k_1\, r + k_2\, r^2.        
\end{eqnarray*}
Notice that all the three models are in the polynomial approximation category. 

\begin{figure*}[htb]
\centering
\includegraphics[width=1.0\textwidth]{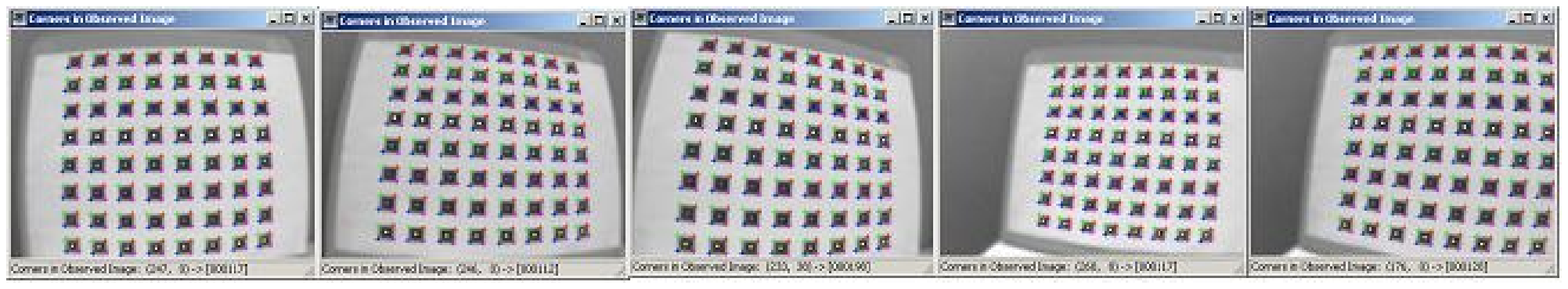}
\caption {Five images of the model plane with the extracted corners (indicated by cross) for the desktop camera.}
\label{fig: extracted desktop}
\end{figure*}

\begin{figure*}[htb]
\centering
\includegraphics[width=1.0\textwidth]{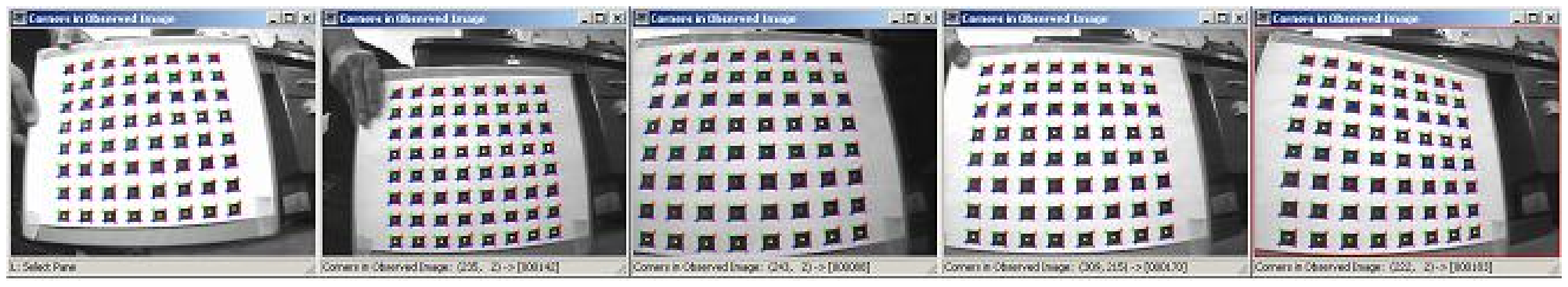}
\caption {Five images of the model plane with the extracted corners (indicated by cross) for the ODIS camera.}
\label{fig: extracted ODIS}
\end{figure*}

Using the public domain test images \cite{zhang98calibrationwebpage}, the desktop camera images \cite{Lilicalreport02} (a color camera in our CSOIS), and the ODIS camera images \cite{Lilicalreport02,odiscamera} (the camera on ODIS robot built in our CSOIS), the final objective function $J$, the 5 estimated intrinsic parameters ($\alpha, \beta, \gamma, u_0, v_0$), and the estimated distortion coefficients ($k_1, k_2$) are shown in Tables \ref{table: Comparison of Three Distortion Models1}, \ref{table: Comparison of Three Distortion Models2}, and \ref{table: Comparison of Three Distortion Models3} respectively. The extracted corners for the model plane of the desktop and the ODIS cameras are shown in Figs.~\ref{fig: extracted desktop} and \ref{fig: extracted ODIS}. As noticed from these images, the two cameras both experience a barrel distortion. The plotted dots in the center of each square are only used for judging the correspondence with the world reference points. 

From Tables \ref{table: Comparison of Three Distortion Models1}, \ref{table: Comparison of Three Distortion Models2}, and \ref{table: Comparison of Three Distortion Models3}, it is observed that the final value of $J$ of {\tt model$_3$} is always greater than that of {\tt model$_1$}, but much smaller than that of {\tt model$_2$}. 
The comparison between {\tt model$_2$} and {\tt model$_3$} might not be fair since the new model has one more coefficient and it is evident that each additional coefficient in the model tends to decrease the fitting residual. However, our main point is to emphasize that by adding a quadratic term in $\delta_r$, higher accuracy can be achieved without sacrificing the property of having analytical undistortion function. 

A second look at the results reveals that for the camera used in \cite{zhang98calibrationwebpage}, which has a small lens distortion, the advantage of {\tt model$_3$} over {\tt model$_2$} is not so significant. However, when the cameras are experiencing a severe distortion, the radial distortion {\tt model$_3$} gives a much better performance over {\tt model$_2$}, as can be seen from Tables \ref{table: Comparison of Three Distortion Models2} and \ref{table: Comparison of Three Distortion Models3}.

\begin{remark}
Classical criteria that are used in the computer vision to assess the accuracy of calibration includes the radial distortion as one part inherently \cite{Juyang92distortionmodel}. However, to our best knowledge, there is not a systematically quantitative and universally accepted criterion in the literature for performance comparisons among different radial distortion models. Due to this lack of criterion, in our work, the comparison is based on, but not restricted to, the fitting residual of the full-scale nonlinear optimization in (\ref{eqn: objective function}). 
\end{remark}

\begin{remark}
To make the results in this paper reproducible by other researchers for further investigation, we present the options we use for the nonlinear optimization: \texttt{options = optimset(`Display', `iter', `LargeScale', `off', `MaxFunEvals', 8000, `TolX', $10^{-5}$,  `TolFun', $10^{-5}$, `MaxIter', 120)}. The raw data of the extracted feature locations in the image plane are also available \cite{Lilicalreport02}. 
\end{remark}

{
\begin{table}[htb]
\centering
\caption{Comparison of Distortion Models Using Public Images}
\label{table: Comparison of Three Distortion Models1}
\renewcommand{\arraystretch}{1}
\vspace{-2mm}
{\begin {tabular}{|c|r|r|r|}\hline
\multicolumn{4}{|c|}{\bf Public Images}\\\hline
{\bf Model} & $\#1$  & $\#2$ & $\#3$ \\\hline
{\bf $J$}& 144.8802 & 148.2789 & 145.6592 \\\hline
$\alpha$ & 832.4860 & 830.7425 & 833.6508 \\
$\gamma$ &   0.2042 &   0.2166 &   0.2075 \\
$u_0$    & 303.9605 & 303.9486 & 303.9847 \\
$\beta$  & 832.5157 & 830.7983 & 833.6866 \\
$v_0$    & 206.5811 & 206.5574 & 206.5553 \\
$k_1$    &  -0.2286 &  -0.1984 &  -0.0215 \\
$k_2$    &   0.1905 &        - &  -0.1566 \\\hline
\end {tabular}}
\end{table}}

{
\begin{table}[htb]
\centering
\caption{Comparison of Distortion Models Using Desktop Images}
\label{table: Comparison of Three Distortion Models2}
\renewcommand{\arraystretch}{1}
\vspace{-2mm}
{\begin {tabular}{|c|r|r|r|}\hline
\multicolumn{4}{|c|}{\bf Desktop Images}\\\hline
{\bf Model} & $\# 1$    & $\# 2$ & $\# 3$ \\\hline
{\bf $J$}& 778.9767 &  904.6797 & 803.3074   \\\hline
$\alpha$ & 277.1449 &  275.5953 & 282.5642   \\      
$\gamma$ &  -0.5731 &   -0.6665 &  -0.6199   \\      
$u_0$    & 153.9882 &  158.2016 & 154.4913   \\      
$\beta$  & 270.5582 &  269.2301 & 275.9019   \\      
$v_0$    & 119.8105 &  121.5257 & 120.0924   \\      
$k_1$    &  -0.3435 &   -0.2765 &  -0.1067   \\      
$k_2$    &   0.1232 &         - &  -0.1577   \\\hline
\end {tabular}}
\end{table}}

{
\begin{table}[htb]
\centering
\caption{Comparison of Distortion Models Using ODIS Images}
\label{table: Comparison of Three Distortion Models3}
\renewcommand{\arraystretch}{1}
\vspace{-2mm}
{\begin {tabular}{|c|r|r|r|}\hline
\multicolumn{4}{|c|}{\bf ODIS Images} \\\hline
{\bf Model} & $\# 1$ & $\# 2$ & $\# 3$ \\\hline
{\bf $J$}& 840.2650 & 933.0981 & 851.2619 \\\hline
$\alpha$ & 260.7658 & 258.3193 & 266.0850 \\      
$\gamma$ &  -0.2741 &  -0.5165 &  -0.3677 \\      
$u_0$    & 140.0581 & 137.2150 & 139.9198 \\      
$\beta$  & 255.1489 & 252.6856 & 260.3133 \\      
$v_0$    & 113.1727 & 115.9302 & 113.2412 \\      
$k_1$    &  -0.3554 &  -0.2752 &  -0.1192 \\      
$k_2$    &   0.1633 &        - &  -0.1365 \\\hline
\end {tabular}}
\end{table}}

\thispagestyle{empty}
\section{Concluding Remarks}
\label{sec: Conclusion}


This paper proposes a new radial distortion model for camera calibration that belongs to the polynomial approximation category. The appealing part of this distortion model is that it preserves high accuracy together with an easy analytical undistortion formula. Performance comparisons are made between this new model with two other existing polynomial radial distortion models. Experiments results are presented showing that this distortion model is quite accurate and efficient especially when the actual distortion is significant. 

\bibliography{calibration,csois1,csois2}
\end{document}